\documentclass[11pt]{article}

\PassOptionsToPackage{table}{xcolor}

\usepackage{enumitem}
\usepackage{times}
\usepackage{latexsym}
\usepackage{booktabs}
\usepackage[T1]{fontenc}
\usepackage[utf8]{inputenc}
\usepackage{microtype}
\usepackage{inconsolata}
\usepackage{graphicx}
\usepackage{tcolorbox}
\tcbuselibrary{skins}
\usepackage{tikz}
\usepackage{float}
\usepackage{multirow}
\usepackage[normalem]{ulem}
\usepackage[final]{acl}
\usetikzlibrary{positioning, fit, arrows.meta, backgrounds}

\definecolor{teal50}{RGB}{225, 245, 238}
\definecolor{teal600}{RGB}{15, 110, 86}
\definecolor{teal800}{RGB}{8, 80, 65}
\definecolor{coral50}{RGB}{250, 236, 231}
\definecolor{coral600}{RGB}{153, 60, 29}
\definecolor{coral800}{RGB}{113, 43, 19}
\definecolor{purple50}{RGB}{238, 237, 254}
\definecolor{purple600}{RGB}{83, 74, 183}
\definecolor{purple800}{RGB}{60, 52, 137}
\definecolor{gray50}{RGB}{241, 239, 232}
\definecolor{gray600}{RGB}{95, 94, 90}

\tikzset{
  every node/.style={font=\scriptsize},
  inputbox/.style={
    draw=gray600, fill=gray50,
    rounded corners=3pt,
    text width=3.8cm,
    align=center,
    inner sep=4pt,
    line width=0.4pt
  },
  llmbox/.style={
    draw=purple600, fill=purple50,
    rounded corners=3pt,
    text width=2.2cm,
    align=center,
    inner sep=3pt,
    line width=0.4pt
  },
  outerbox/.style={
    rounded corners=4pt,
    inner sep=4pt,
    line width=0.4pt
  },
  subbox/.style={
    rounded corners=2pt,
    text width=2.6cm,
    align=center,
    inner sep=3pt,
    line width=0.4pt,
    minimum height=0.9cm
  },
  tealouter/.style={outerbox, draw=teal600, fill=teal50},
  coralouter/.style={outerbox, draw=coral600, fill=coral50},
  tealsub/.style={subbox, draw=teal600, fill=teal50!55},
  coralsub/.style={subbox, draw=coral600, fill=coral50!55},
  arr/.style={
    -{Stealth[length=3.5pt, width=2.5pt]},
    line width=0.4pt,
    color=gray600
  }
}

\title{Scene Abstraction for Lexical Semantics: \\
Structured Representations of Situated Meaning}

\author{Yejin Cho \\
  The University of Texas at Austin \\
  Department of Linguistics \\
  \texttt{ycho@utexas.edu} \\\And
  Katrin Erk \\
  University of Massachusetts, Amherst \\
  Department of Linguistics \\
  \texttt{kerk@umass.edu} \\}

\begin{document}
\maketitle

\begin{abstract}
\textit{Coffee} and \textit{tea} share many properties, yet
they evoke strikingly different situations, atmospheres, and
affective associations. These situated dimensions of word
meaning are real and systematic, but they remain implicit in
most computational representations of lexical meaning. 
We propose \textit{Scene Abstraction}, a framework for
constructing structured representations of the interpretive
scenes that words participate in across usage contexts.
Each scene consists of a \textit{Contextual Scene}
(Events, Entities, Setting) and an expression-centered
\textit{Expression Profile} (Engaged events, Generalizable
properties, Evoked emotions), operationalized through
few-shot prompting of a large language model.
Our contributions are threefold:
(1) a structured representation framework for situated lexical
meaning;
(2) COCA-Scenes, a dataset of 520 usage instances across
26 keywords for distinct scene identification; and
(3) empirical evidence from two experiments suggesting
that scenes are reliably identifiable across human observers 
(82.4\% accuracy, +11.8 pp over text-only embeddings) 
and that our scene profiles more closely align with human interpretation
of words in context than ATOMIC-based alternatives
(86.4\% preference across three semantic dimensions).
\end{abstract}

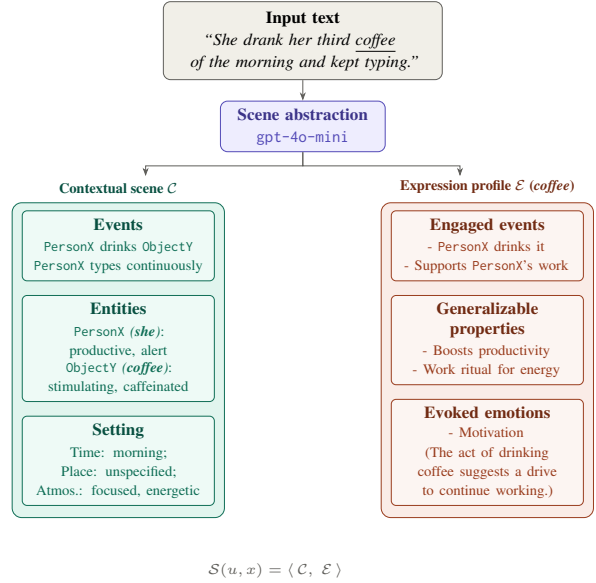
\begin{figure}[t]
\centering
\resizebox{\columnwidth}{!}{%
\begin{tikzpicture}[node distance=0.3cm]
\node[inputbox] (input) {%
  \textbf{Input text}\\[1pt]
  \textit{``She drank her third \underline{coffee}}\\
  \textit{of the morning and kept typing.''}%
};
\node[llmbox, below=0.3cm of input] (llm) {%
  \textcolor{purple800}{\textbf{Scene abstraction}}\\[1pt]
  \textcolor{purple600}{\texttt{gpt-4o-mini}}%
};
\draw[arr] (input.south) -- (llm.north);
\draw[arr] (llm.south) -- ++(0,-0.2) -| ++(-2.3,0) -- ++(0,-0.15);  
\draw[arr] (llm.south) -- ++(0,-0.2) -|  ++(2.3,0) -- ++(0,-0.15);  
\node[tealsub, below=0.85cm of llm, xshift=-2.7cm] (events) {%
  \textcolor{teal800}{\textbf{Events}}\\[1pt]
  \textcolor{teal600}{\tiny\texttt{PersonX} drinks \texttt{ObjectY}}\\
  \textcolor{teal600}{\tiny\texttt{PersonX} types continuously}%
};
\node[tealsub, below=0.2cm of events] (entities) {%
  \textcolor{teal800}{\textbf{Entities}}\\[1pt]
  \textcolor{teal600}{\tiny\texttt{PersonX} \textit{(\textbf{she})}:\ productive, alert}\\
  \textcolor{teal600}{\tiny\texttt{ObjectY} \textit{(\textbf{coffee})}:\ \\stimulating, caffeinated}%
};
\node[tealsub, below=0.2cm of entities] (setting) {%
  \textcolor{teal800}{\textbf{Setting}}\\[1pt]
  \textcolor{teal600}{\tiny Time: morning;}\\
  \textcolor{teal600}{\tiny Place: unspecified;}\\
  \textcolor{teal600}{\tiny Atmos.: focused, energetic}%
};
\begin{scope}[on background layer]
\node[tealouter,
  fit=(events)(entities)(setting),
  label={[teal800, font=\tiny\bfseries]above:Contextual scene $\mathcal{C}$},
  yshift=-0.03cm
] (gc) {};
\end{scope}
\node[coralsub, below=0.85cm of llm, xshift=2.7cm] (engev) {%
  \textcolor{coral800}{\textbf{Engaged events}}\\[1pt]
  \textcolor{coral600}{\tiny- \texttt{PersonX} drinks it}\\
  \textcolor{coral600}{\tiny- Supports \texttt{PersonX}'s work}%
};
\node[coralsub, below=0.2cm of engev] (props) {%
  \textcolor{coral800}{\textbf{Generalizable properties}}\\[1pt]
  \textcolor{coral600}{\tiny - Boosts productivity}\\
  \textcolor{coral600}{\tiny - Work ritual for energy}%
};
\node[coralsub, below=0.2cm of props] (emo) {%
  \textcolor{coral800}{\textbf{Evoked emotions}}\\[1pt]
  \textcolor{coral600}{\tiny - Motivation \\
  (The act of drinking coffee suggests a drive to continue working.)}%
};
\begin{scope}[on background layer]
\node[coralouter,
  fit=(engev)(props)(emo),
  label={[coral800, font=\tiny\bfseries]above:Expression profile $\mathcal{E}$ (\textit{coffee})},
  yshift=-0.03cm
] (ep) {};
\end{scope}
\node[below=0.7cm of setting, xshift=2.3cm,
  font=\tiny, text=gray600, align=center] {%
  $\mathcal{S}(u,x) = \langle\,\mathcal{C},\;\mathcal{E}\,\rangle$%
};
\end{tikzpicture}%
}
\caption{%
  The Scene Abstraction framework. Given a usage context \textit{u}
  and target expression \textit{x} (\textit{coffee}) in it, an LLM
  produces $\mathcal{S}(u,x)$ comprising a \textbf{Contextual
  Scene} $\mathcal{C}$ (\textit{Events, Entities, Setting}) that captures
  the overall situation described by \textit{u}, and an
  \textbf{Expression Profile} $\mathcal{E}$ (\textit{Engaged events,
  Generalizable properties, Evoked emotions}) that characterizes
  the scene-grounded meaning of \textit{x}.%
}
\label{fig:framework}
\end{figure}

\section{Introduction}
What does \textit{crow} mean? The dictionary answer---a large
black bird---captures little of how the word actually functions
in context. In practice, \textit{crow} participates in scenes
of silence, isolation, ominous atmospheres, winter landscapes, and death symbolism;
these recurring situational patterns are part of what it means
to know the word, yet none are explicitly encoded in its
dictionary definition. Similarly, words such as \textit{whiskey},
\textit{rain}, or \textit{rose} evoke recurring experiential
and affective situations that extend well beyond referential
content. These situated dimensions of meaning are not incidental;
they reflect structured interpretive regularities that shape how
words are understood in context.

Capturing this situated, interpretive character of meaning
computationally, however, remains an open challenge.
Frame semantics \citep{fillmore1976frame,fillmore1982frame} proposes that
word meaning is grounded in rich background knowledge,
including situational, affective, and cultural dimensions.
The primary large-scale implementation of this tradition,
FrameNet \citep{baker1998berkeley}, however, focuses on
a narrower set of semantic frames describing
predicate-argument structure. It does not readily scale
to the open-ended, contextually variable dimensions of
situated meaning, such as \textit{crow} being ominous in
one scene and social or intelligent in another.
Distributional models
\citep{harris1954distributional, landauer1997solution} and their
contextual successors \citep{peters2018deep, devlin2019bert}
capture rich usage-based patterns and scale to large corpora,
but encode semantic structure---events, participant roles,
atmosphere, affective associations---implicitly within dense
vectors that are difficult to inspect, compare, or interpret
directly. 
Instruction-tuned LLMs demonstrate strong contextual
understanding and can produce nuanced interpretations of words
in context, but their default outputs are free-form prose---unstructured,
variable across prompting conditions, and difficult to compare
or aggregate systematically across usage instances.
The framework we propose draws on all three traditions,
operationalizing the situated, affective, and cultural
dimensions of word meaning that frame semantics foregrounds, 
with the scalability of distributional approaches
and the generative capacity of LLMs.

We propose \textit{Scene Abstraction}, a framework that views
lexical meaning as structured distributions over interpretive
scenes $\mathcal{S}(u, x)$, where $u$ is a usage context and
$x$ is a target expression. 
The central representational idea is that some aspect of word meaning in context
can be more effectively captured by the structured
scene it instantiates than by its surface textual surroundings
alone.
This framework attempts to understand aspects of lexical meaning through the characteristic patterns of scenes across usage instances of a word.
As illustrated in Figure~\ref{fig:framework}, each scene
consists of two complementary components: a \textit{Contextual
Scene} $\mathcal{C}$ capturing the broader situational
interpretation---events, entities, and setting---and, crucially,
a target-expression-centered \textit{Expression Profile}
$\mathcal{E}$ capturing the scene-grounded meaning of $x$
specifically: the events it engages in, the properties it
bears within the scene, and the affective associations it
evokes.
Different words in the same sentence will have
different Expression Profiles:
it is the
Expression Profile that makes this per-expression meaning
realization explicit and inspectable. Scenes are not
predefined ontological categories but open-ended interpretive
representations constructed dynamically for each usage
instance.

We evaluate the framework through two experiments addressing
distinct validation goals. First, an \textit{odd-scene-out}
task assesses construct validity:
do scene-level distinctions constitute a reliably shared discriminative structure among humans, and 
do automatically generated scene-based representations outperform raw-text embeddings in capturing them?
Second, a human preference study evaluates
output quality: do LLM-generated scene abstractions align with
human interpretation of words in context? Specifically, we compare them
against ATOMIC-based commonsense profiles \citep{sap2019atomic,
hwang2021comet}, which represent the closest existing
alternative for structured situational inference.

This paper makes the following contributions:

\begin{itemize}[leftmargin=*, itemsep=2pt]
    \item We introduce \textbf{Scene Abstraction}, a two-layer
    framework comprising a Contextual Scene $\mathcal{C}$ (events, entities, setting) and a target-expression-centered Expression Profile $\mathcal{E}$, operationalized via few-shot prompting of a large language model.

    \item We construct \textbf{COCA-Scenes}, a dataset of 520
    usage instances across 26 keywords designed to support
    scene-level evaluation.

    \item We provide empirical evidence that scene-based
    representations outperform text-only embeddings in
    human-aligned situational discrimination (69.3\% vs.\
    57.5\% accuracy), and that LLM-generated scene abstractions
    align substantially more closely with human interpretation of words in context
    than ATOMIC-base alternatives (86.4\% preference).
\end{itemize}

\section{Related Work}
\label{sec:related}
Existing approaches to lexical meaning representation, characterization of situations, and computational frameworks
for commonsense knowledge offer important insights on which the current work builds.

\paragraph{Frame Semantics.}
Frame Semantics \citep{fillmore1976frame, fillmore1982frame}
proposed that understanding a word requires access to rich background knowledge about the situations, participants, and relations it evokes. Construction grammar has similarly argued that meaning is inseparable from structured situational knowledge \citep{goldberg1995constructions}.
Scene Abstraction builds on this foundational insight, extending it computationally through the use of LLMs to construct open-ended, usage-grounded scene representations for any expression and context.
The primary large-scale implementation of this tradition, FrameNet \citep{baker1998berkeley}, focuses on a particular facet of Frame Semantics: semantic frames describing the core meaning and predicate-argument structure.
Scene Abstraction, by contrast, requires no predefined inventory for interpretation, allowing it to flexibly capture how the same word takes on different meanings depending on context, such as \textit{crow} being ominous in one scene and intelligent or social in another.

\paragraph{Situation Models, Scripts, and Scene.}
Research on discourse comprehension has established that language
understanding involves constructing rich mental representations
beyond surface linguistic form \citep{vandijk1983strategies,
zwaan1995dimensions, zwaan1998situation}. Script-based approaches
further emphasize stereotyped event sequences in organizing
situational knowledge \citep{schank1977scripts}, and such
structures have been induced from large text corpora
\citep{chambers2008unsupervised, pichotta2016using}. Scene
Abstraction draws on this tradition but differs in two respects.
Unlike situation models, which remain implicit cognitive
representations, scene abstraction \textit{externalizes} them
in explicit, structured, and computationally accessible form.
Unlike scripts, which are organized around temporal event
sequences, 
a scene is closer to a \textit{situational
snapshot}: the structured interpretation formed at a given
moment, in which events, participants, atmosphere, and affect
are jointly organized within a single instance.
The scene abstraction approach prioritizes this within-moment
interpretive variability over the sequential structure that
scripts encode.

\paragraph{Distributional and Contextual Semantics.}
Distributional and contextual embedding models
\citep{harris1954distributional, lee2018feel, peters2018deep, devlin2019bert, wiedemann2019does, schlechtweg2020semeval} offer strong
computability, but encode semantic structure implicitly within
dense vectors that do not directly expose the event structures,
participant relations, atmosphere, or affective dimensions
important for situated lexical interpretation. \citet{hoyle2023natural}
show that externalizing implicit communicative content via
LLM-generated propositions yields better text representations
across downstream tasks. Scene Abstraction is motivated by a
similar observation but targets specific situated dimensions of
lexical meaning---events, participants, atmosphere, and
expression-specific profiles---organized into a fixed, reusable
schema for cross-instance comparison. Because each scene
component can be independently encoded as a text embedding, the
resulting distributions are both interpretable and computable.

\paragraph{Commonsense and Event Knowledge Representation.}
Large-scale commonsense knowledge graphs \citep{sap2019atomic,
bosselut2019comet, hwang2021comet} and narrative-grounded causal
knowledge resources \citep{mostafazadeh2020glucose} have
substantially advanced the representation of implicit world
knowledge---causal relations, social dynamics, and event
structure. These resources are primarily designed to support
general-purpose commonsense inference and narrative understanding,
however, rather than the representation of lexical meaning
specifically. Their relational schemas describe what typically
holds in everyday situations in general, and do not directly
address how the meaning of a particular word is shaped by the
specific context in which it appears. Scene Abstraction shares
the goal of making implicit situational knowledge explicit, but
focuses specifically on what a target expression does, implies,
and evokes in a given usage instance---a question that
commonsense knowledge resources are not designed to answer.
This distinction is directly tested in Experiment~2, where
ATOMIC-based profiles consistently produce outputs less aligned
with human scene-grounded interpretation of target words
than Scene Profiles.

\paragraph{LLMs for Structured Semantic Generation.}
LLMs have demonstrated strong capabilities for generating
structured semantic representations \citep{brown2020language,
wei2022chain, west2022symbolic, gilardi2023chatgpt, wang2023chatgpt}. 
Closely related to our work, \citet{gu2022dream} shows that explicitly elaborating situational context before answering improves downstream QA performance, providing evidence that structured
situational elaboration is a productive intermediary step for
language understanding. 
Scene Abstraction builds on these capabilities by supplying
a fixed representational schema that channels LLM interpretive capacity into 
more structured and explicit scene representations.

\smallskip
\smallskip
\smallskip
\noindent Scene Abstraction inherits from Frame Semantics the core idea that word meaning is grounded in rich background knowledge, including situational, affective, and cultural dimensions. It draws further on the situational grounding of discourse comprehension research and the scalability of distributional approaches. 
What it contributes is a computational method for operationalizing this background knowledge at the level of individual usage instances: 
based on a manually designed scene schema, we use LLMs to construct structured, easily interpreted scene representations for any word in any context with great flexibility.

\section{Scene Abstraction Framework}
\label{sec:framework}

\subsection{Motivation}

A central observation motivating our framework is that word
meaning in context is not simply a function of local lexical
co-occurrence. When readers encounter a sentence such as:

\begin{quote}
\textit{The man sat alone at the kitchen table, drinking whiskey
late at night.}
\end{quote}

\noindent their interpretation of \textit{whiskey} draws on a
structured understanding of the entire situation: the solitude,
the hour, the setting, the emotional weight implied by the scene.
This interpretive structure---events, participants, atmosphere,
and the target expression's role within it---is central to how
the word is understood in context.
A man drinking whiskey alone at night and a man drinking beer with friends at a pub involve the same surface event---drinking an alcoholic beverage---yet give rise to substantially different scenes.
Distinguishing these cases may require a representation that
encodes more than just the surface event.

We formalize this observation through the Scene Abstraction
framework: the contextual meaning of $x$ in $u$ can be 
more effectively captured by a structured scene representation than by surface
textual context alone, and type-level lexical meaning can be
understood as a distribution of such scenes across usage instances.

\begin{table}[t]
\centering
\small
\renewcommand{\arraystretch}{1.2}
\begin{tabular}{p{2.2cm} p{4.8cm}}
\toprule
\textbf{Component} & \textbf{Content} \\
\midrule
\multicolumn{2}{l}{\textit{Contextual Scene $\mathcal{C}$}} \\
\midrule
Events &
  \texttt{PersonX sits at ObjectY} \newline
  \texttt{PersonX drinks ObjectZ} \newline
  \texttt{PersonX is alone at ObjectY} \\
\addlinespace
Entities &
  \texttt{PersonX} \textit{(the man)}: solitary, melancholy; \texttt{ObjectZ} \textit{(whiskey)}: alcoholic, comforting \\
\addlinespace
Setting &
  Kitchen; late at night; \newline
  somber and reflective \\
\midrule
\multicolumn{2}{l}{\textit{Expression Profile $\mathcal{E}$}
(\textit{whiskey})} \\
\midrule
Engaged Events &
  \texttt{PersonX drinks it} \newline
  \texttt{It is consumed alone at ObjectY} \\
\addlinespace
Generalizable Properties &
  Often associated with solitude and reflection; can signify coping mechanisms during difficult times \\
\addlinespace
Evoked Emotions &
  Melancholy; loneliness \\
\bottomrule
\end{tabular}
\caption{Scene output (abbreviated) for \textit{whiskey} in \textit{``The
man sat alone at the kitchen table, drinking whiskey late at
night.''}, generated by \texttt{gpt-4o-mini}. The
Contextual Scene $\mathcal{C}$ captures the broader situational
interpretation; the Expression Profile $\mathcal{E}$ captures
the scene-grounded meaning of the target expression.}
\label{tab:scene_example}
\end{table}

\subsection{Formal Definitions}

\begin{description}[leftmargin=1em, itemsep=3pt]
    \item[Usage context] A natural language context $u$ in
    which a target expression $x$ occurs.
    \item[Usage instance] A pair $(u, x)$ where $u$ is a
    usage context and $x$ is a target expression
    (a word or multi-word expression) occurring in $u$.
    \item[Scene representation] A structured interpretive representation
    $\mathcal{S}(u, x)$ constructed for usage instance
    $(u, x)$, consisting of two complementary components: a
    \textit{Contextual Scene} $\mathcal{C}$ capturing the
    broader situational interpretation of $u$, and an
    \textit{Expression Profile} $\mathcal{E}$ capturing the
    scene-grounded meaning of $x$ within that context.
    \item[Scene set] Given a set of usage contexts $\{u_i\}_{i=1}^{N}$ containing $x$, the corresponding set of scene representations $\{\mathcal{S}(u_i, x)\}_{i=1}^{N}$ constitutes the \textit{scene set} of $x$ with respect to that corpus. Type-level properties of $x$ can then be studied through the patterns observed across this collection.
\end{description}

\subsection{Contextual Scene}

The Contextual Scene $\mathcal{C}$ represents the broader situational interpretation of a usage instance, independent of which specific expression is targeted. It consists of three components: \textit{Events} (abstracted actions and interactions, expressed using anonymized participant labels such as \texttt{PersonX}, \texttt{ObjectX}), \textit{Entities} (participants and objects salient to the scene, characterized by their role with frame, properties, and emotional state), and \textit{Setting} (spatial, temporal, and atmospheric background). See Table~\ref{tab:scene_example} for an example.

While the empirical evaluation in this paper
focuses entirely on the Expression Profile, the Contextual
Scene is included as part of the full scene representation
for its utility in downstream analyses.

\subsection{Expression Profile}

The Expression Profile $\mathcal{E}$ captures how the target
expression $x$ is interpreted within the surrounding scene.
Different words in the same sentence will have different Expression Profiles: 
what \textit{whiskey} engages in, what properties it bears, and what emotions it evokes in the kitchen scene differs substantially from what \textit{table} does in the same context. It consists of three
sub-components.

\paragraph{Engaged Events.}
Events in which $x$ plays a central role within the scene.

\paragraph{Generalizable Properties.}
Semantic properties associated with $x$ in the scene context.

\paragraph{Evoked Emotions.}
Affective associations evoked by $x$ in the scene context.

\subsection{Worked Example}

We describe the full LLM-based generation procedure in
Section~\ref{sec:implementation}. Table~\ref{tab:scene_example}
presents a complete scene abstraction for the sentence
\textit{``The man sat alone at the kitchen table, drinking
whiskey late at night,''} with \textit{whiskey} as the target
expression, generated by \texttt{gpt-4o-mini}.

\subsection{Scene Similarity}

Scene Abstraction comes with a natural notion of scene similarity: given two usage instances $u_i$ and $u_j$, 
the similarity between $\mathcal{S}(u_i, x)$ and $\mathcal{S}(u_j, x)$ provides a structured, interpretable basis for comparing how $x$ is used across contexts.
Each scene representation can be mapped to a dense vector
by embedding its components using a pretrained sentence
encoder. This can be done either by computing a single
overall embedding across all components, or by embedding
individual components separately, allowing selective focus
on specific dimensions of the scene such as events,
properties, or evoked emotions.
Given a scene set $\{\mathcal{S}(u_i, x)\}_{i=1}^{N}$ for a word $x$, variation at scene level across contexts can be inspected, making the approach applicable to tasks such as the study of polysemy, contextual variation, sociolinguistic meaning differences, and lexical semantic change.

\section{Implementation}
\label{sec:implementation}

\subsection{Overview}

Given a usage instance $(u, x)$, our pipeline produces a
structured scene representation $\mathcal{S}(u, x)$ in two
stages: (1) \textit{scene generation}, in which an LLM
produces a structured abstraction from the usage context via
few-shot prompting; and (2) \textit{scene embedding}, in which
selected components of the structured output are converted into
dense vector representations for downstream comparison.
Figure~\ref{fig:framework} illustrates the scene abstraction process
\footnote{We use \texttt{gpt-4o-mini} for all scene 
generation experiments (temperature = 0.2,
max\_tokens = 512, top\_p = 1, frequency\_penalty = 0,
presence\_penalty = 0).}

\subsection{Scene Generation via Structured Prompting}
\label{sec:scene_generation}

\paragraph{Prompt Design.}
We prompt the model with a system instruction that specifies the desired output format, followed by $k$ few-shot examples that show the intended level of abstraction, and the input sentence with the target expression highlighted.

Figure~\ref{fig:prompt_instruction} shows the full prompt instruction used for scene abstraction.

The instruction is governed by four abstraction principles.
These principles are not independent constraints but work in
a mutually complementary fashion: the first two govern
structural consistency across scenes, while the latter two
ensure that each representation retains sufficient information
for meaningful semantic interpretation.

\begin{description}[leftmargin=1em, itemsep=4pt]
    \item[Generalization.] Proper names and specific referring
    expressions are replaced with role-based labels
    (\texttt{PersonX}, \texttt{PersonY}, \texttt{ObjectX},
    \texttt{AnimalX}, \texttt{PlaceX}), enabling structural
    similarities to be captured even when the same scene
    pattern is realized across different surface forms. This
    generalization is a prerequisite for analyzing recurring
    scene structures beyond surface-level variation.

    \item[Detail Omission.] 
    Narrative details not relevant to the situational interpretation are omitted. For example, \textit{``pulled himself off the wooden floor''} is abstracted to \texttt{PersonX pulls himself off the floor}, dropping incidental details such as the material of the floor.

    \item[Interpretability.]
    All output fields are expressed in natural language that humans can easily read and analyze. For example, entity properties are written as phrases such as \textit{``resilient''} or \textit{``hopeful yet struggling''} rather than numeric identifiers or ontology codes.

    \item[Context Sensitivity.] The Expression Profile is described in terms specific to the given usage context rather than the target expression in general. For example, \textit{stab} in \textit{``Your deed was another stab, piercing my very heart''} yields properties such as \textit{``symbolizes emotional pain or betrayal''}, while \textit{stab} in \textit{``my stab at this version of normality''} yields \textit{``signifies a bold attempt at change''}. Generic descriptions applicable to any context are explicitly discouraged.
\end{description}

\paragraph{Few-Shot Examples.}
We use $k = 3$ few-shot examples covering keywords with
different situational configurations.
Examples were manually constructed to demonstrate the intended level of abstraction. Table~\ref{tab:json_example} presents one such example.

\paragraph{Output Format}
The model is instructed to produce a structured output in JSON format covering the components described in Section~\ref{sec:framework}. Table~\ref{tab:scene_example} shows a representative example of the output format.

\subsection{Scene Embedding}
\label{sec:conditions}
To support vector-space operations over scene representations,
we convert structured scene outputs into dense vector
representations via a two-step procedure. Because scene
representations are structured into distinct components,
any subset---whether from the Contextual Scene or the
Expression Profile---can be independently encoded,
allowing selective focus on the dimensions of
meaning most relevant to a given analysis task.
This modularity may offer an interpretability advantage: different
components can be compared, aggregated, or weighted
independently without retraining.

In the experiments reported here, we focus on the Expression Profile, which
captures the scene-grounded meaning of the target expression
directly and is therefore most amenable to instance-level
semantic comparison.

\begin{table*}[t]
\centering
\small
\renewcommand{\arraystretch}{1.15}
\begin{tabular}{p{3.8cm} p{11.5cm}}
\toprule
\textbf{Few-shot example input} & \textbf{Few-shot example output} \\
\midrule
\textit{``Sometimes she would just stay by the window, feeding
the \textbf{crows} while he was doing some paperwork, just like
old times.''} \newline\newline Keyword: \textit{crow} &
\textbf{Contextual Scene} \newline
\textbf{* Events:} \newline
PersonX stays near a window; PersonX
feeds AnimalGroupX; PersonY does paperwork nearby; PersonX and PersonY share a quiet routine \newline\newline
\textbf{* Entities:} \newline
$\cdot$ PersonX (\textit{she}): Agent (Feeding), Experiencer
(Remembering\_past); Property: Reflective; Emotion: Calm,
Nostalgia \newline
$\cdot$ PersonY (\textit{he}): Co-participant (Routine\_activity),
Worker (Office\_work) \newline
$\cdot$ AnimalGroupX (\textit{the crows}): Recipients (Feeding), Symbols (Memory\_triggering); Property: Accustomed to being fed \newline\newline
\textbf{* Setting:}\newline
$\cdot$ Place: by a window (indoors); Time: reflective
moment; \newline
$\cdot$ Atmosphere: calm and nostalgic \newline\newline
\textbf{Expression Profile} (crow = AnimalGroupX) \newline
$\cdot$ \textit{Engaged events:} PersonX feeds them; they receive
food as part of a habitual routine \newline
$\cdot$ \textit{Generalizable properties:} commonly present near humans; respond to routine interactions; may evoke
memories through repeated presence; passive figures in
quiet domestic routines \newline
$\cdot$ \textit{Evoked emotions:} Nostalgia (tied to memories of past
shared routines); Serenity (presence contributes to a
peaceful atmosphere) \\
\bottomrule
\end{tabular}
\caption{In-context learning example for \textit{crow}. The
output illustrates how the four abstraction principles operate
in practice: role-based labels (\texttt{AnimalGroupX},
\texttt{PersonX}) realize generalization; atmosphere and
evoked emotions are inferred rather than stated; and the
Expression Profile captures situated rather than dictionary
meaning.}
\label{tab:json_example}
\end{table*}

\paragraph{Textual Serialization.}
Selected scene components are serialized into natural language
strings. For example, the \texttt{evoked\_emotions} field
\texttt{["loneliness", "resignation", "introspection"]} is
serialized as \textit{``evoked emotions: loneliness,
resignation, introspection.''} 

\paragraph{Embedding.}
Serialized strings are encoded using
SentenceBERT (\texttt{all-mpnet-base-v2}; \citet{reimers2019sentence}).
For
the text-only baseline, raw input text is encoded using
the same model to ensure comparability.

\paragraph{Representation Conditions.}
We compare the following representation conditions in
Experiment~1 (Section~\ref{sec:exp1}):

\begin{itemize}[itemsep=1pt]
    \item \textbf{Text}: text embedding of the raw input
    sentence (baseline).
    \item \textbf{Text + Event}: original text concatenated
    with serialized Engaged Events from the expression profile.
    \item \textbf{Text + Property}: original text
    concatenated with serialized Generalizable Properties.
    \item \textbf{Text + Emotion}: original text
    concatenated with serialized Evoked Emotions.
    \item \textbf{Text + Scene}: original text concatenated
    with all three expression profile components.
    \item \textbf{Scene Only}: all three expression profile
    components without the original text.
\end{itemize}

This design allows us to assess both the overall contribution
of the expression profile to downstream task performance and
the relative discriminative weight of each individual
component.

\section{Experiments}

We evaluate Scene Abstraction through two experiments
addressing the following research questions:

\begin{itemize}[itemsep=1pt]
    \item \textbf{RQ1}: Are scene-level distinctions reliably
    shared among human observers, and do scene-based
    representations capture them better than text embeddings?
    \item \textbf{RQ2}: Do LLM-generated scene abstractions
    align with human interpretation of scene-grounded word meaning?
\end{itemize}

The two experiments reported here focus on the
Expression Profile $\mathcal{E}$. The Contextual Scene
$\mathcal{C}$ is retained as background to the Expression
Profile and as a basis for future analyses.

\subsection{Experiment 1: Odd-Scene-Out}
\label{sec:exp1}

\subsubsection{Task and Motivation}

We evaluate scene-based representations through an
\textit{odd-scene-out} task: given five usage instances of
the same target expression, identify the one whose contextual
interpretation is least compatible with the other four.
Scenes have many facets, and different observers may weigh them differently. However, when a group of usage instances shares a common situational pattern, that shared pattern can be reliably identified even without agreement on a category label. The odd-scene-out design exploits this: rather than asking annotators to categorize scenes, it tests whether they can identify the instance that does not share the situational pattern of the other four.

\subsubsection{Dataset: COCA-Scenes}

We construct the \textbf{COCA-Scenes} dataset to support this
experiment.
Twenty-six keywords were manually selected (e.g., \textit{bathroom}, \textit{eagle}, \textit{sword}; 
see Appendix~\ref{app:keywords} for the full list)
based on their potential to evoke vivid and varied situational contexts.
For each keyword, four distinct scene types were defined through iterative inspection of corpus examples, aiming for situationally distinct configurations that yield clearly contrasting scene-level interpretations.
Five sentences per scene type were manually collected from the fiction genre of the Corpus of Contemporary American English
\citep[COCA;][]{davies2008corpus}.
COCA is a large, genre-diverse corpus of American English; its fiction texts were chosen for their rich depictions of events, participants, and atmosphere.

This yielded a total of 520 sentences (26 keywords $\times$ 4 scene types $\times$ 5 sentences).
Each odd-scene-out trial is constructed by sampling
four sentences from one scene type (the ``base'' scenes) and
one sentence from a different scene type for the same keyword
(the ``odd'' scene). Table~\ref{tab:exp1-stimuli} shows
representative stimuli for three keywords.

\begin{table}[t]
\centering
\small
\begin{tabular}{p{1.2cm} p{5.8cm}}
\toprule
\textbf{Keyword} & \textbf{Sentences (odd scene in bold)} \\
\midrule
\textit{raccoon} &
  (a) Ash-can covers clank, and a raccoon makes its furtive
  way down the rain vent. \\
  & (b) \textbf{He always seemed to find something to rescue:
  a raccoon one summer, a squirrel the next.} \\
  & (c) A rabid raccoon ate through the screen and leaped on
  her bed, snarling. \\
  & (d) When raccoons try to get on our porch, Momma chases
  em off with a broom. \\
  & (e) Raccoons pried open the doors and tore into the
  boxes. \\
\addlinespace
\textit{fire} &
  (a) He rocked in the chair, staring at the fire. Both
  brothers were quiet. \\
  & (b) Sarah had fallen asleep next to the fire; Thomas had
  nodded off in his chair. \\
  & (c) With a tiny fire guarding the entrance, Jacques felt
  downright cozy. \\
  & (d) \textbf{The fire was up around his head and the smoke
  was so thick we couldn't see him.} \\
  & (e) Mr.\ O'Flynn sat down to warm his hands by the fire. \\
\addlinespace
\textit{bathroom} &
  (a) He stared at his face in the bathroom mirror, trying
  to see it through his cousins' eyes. \\
  & (b) The bathroom door was open, and shards of glass lay
  over the tiles. \\
  & (c) \textbf{He was found dead in the bathroom at the
  mall, having wandered in and collapsed.} \\
  & (d) She dreamed of accidents---slipping in the tub,
  stabbing her head on the cross in the bathroom. \\
  & (e) In her limp hand is a gun. Speckling the white tile
  of the bathroom is blood. \\
\bottomrule
\end{tabular}
\caption{Representative odd-scene-out stimuli. Each trial
presents five sentences sharing a target keyword; participants
identify the one depicting a situationally distinct scene
(shown in bold).}
\label{tab:exp1-stimuli}
\end{table}

\subsubsection{Human Evaluation}
\label{sec:exp1-human}

Six participants were recruited and divided into two 
independent groups (Group 1, Group 2) of
three annotators each.\footnote{This study received an exempt determination from the Institutional Review Board at the University of Texas at Austin (IRB ID: STUDY00008122).}
Each participant evaluated 52 trials
(13 keywords $\times$ 4 trials per keyword) and selected the
sentence depicting the situationally distinct scene. Every
trial was independently judged by all three members within
each group, enabling calculation of inter-annotator agreement
(IAA).

We report Gwet's AC1 \citep{gwet2008computing} as our primary IAA measure, which estimates chance agreement as a function of how evenly ratings are
distributed across categories rather than from marginal
frequencies alone, making it robust precisely when one
category dominates.
It therefore serves as the primary agreement metric in this study, a decision applied consistently in Experiment~2 as well.

\subsubsection{Computational Evaluation}
\label{sec:exp1-auto}

In parallel with the human study, we assess whether
scene-based representations provide a discriminative signal
for the same task. We use 
SentenceBERT (\texttt{all-mpnet-base-v2}; \citet{reimers2019sentence}) to generate sentence embeddings
for each candidate. The candidate with the lowest mean
pairwise cosine similarity to the remaining four is selected
as the predicted odd-scene-out item. We compare the
representation conditions defined in
Section~\ref{sec:conditions}.

\subsubsection{Results}
\label{sec:exp1_iaa}

\paragraph{Human performance and agreement.}

Human annotators achieved an overall accuracy of
\textbf{82.37\%} against the COCA-Scenes gold standard scene type labels,
far exceeding the chance level of 20\%
(five-way choice). Individual accuracy was stable across
participants (range: 73.08--94.23\%; median: 82.69\%),
indicating that scene-level distinctions are robust enough to
support reliable identification.

Table~\ref{tab:iaa_results} reports IAA. Gwet's AC1 yields a mean score of \textbf{0.761} (\textit{Substantial} agreement, per \citet{landis1977measurement}).
This is consistent with the view that scene-level distinctions are reliably identifiable from shared interpretive structure.

\begin{table}[t]
\centering
\small
\begin{tabular}{lccc}
\toprule
\textbf{Metric} & \textbf{Group 1} & \textbf{Group 2} &
\textbf{Mean} \\
\midrule
Human Accuracy & 87.82\% & 76.92\% & 82.37\% \\
Full Agreement & 78.85\% & 71.15\% & 75.00\% \\
Gwet's AC1     & 0.821 & 0.702 & 0.761 \\
\bottomrule
\end{tabular}
\caption{Human accuracies and inter-annotator agreement for Experiment 1 (Odd-scene-out).} 
\label{tab:iaa_results}
\end{table}

\paragraph{Computational discriminability.}

Table~\ref{tab:automatic_results} reports the accuracy of
model predictions on the odd-scene-out task across input conditions.
The text-only baseline achieves 0.575.
Integrating the complete set of scene features (Event,
Property, Emotion) raises accuracy to \textbf{0.693}, a gain
of +11.8 percentage points. The scene-only condition,
which relies solely on abstracted
scene features without the original sentence, achieves
0.627, surpassing the text-only baseline and demonstrating
that abstracted scene features alone carry sufficient semantic
weight to identify situational outliers.
Among individual scene components, \textit{Property} proves most discriminative (0.661), suggesting that state-based and descriptive attributes are particularly central to defining scene
boundaries for the concrete nouns in this dataset, though this pattern may not generalize to other word classes.

\begin{table}[t]
\centering
\small
\begin{tabular}{lp{3.5cm}c}
\toprule
\textbf{Input} & \textbf{Scene feature} & \textbf{Acc.} \\
\midrule
Text only & --- & 0.575 \\
\midrule
\multirow{4}{*}{Text + Scene}
  & All (Event+Prop+Emo) & \textbf{0.693} \\
  & Event only     & 0.609 \\
  & Property only  & 0.661 \\
  & Emotion only   & 0.562 \\
\midrule
Scene only & All (Event+Prop+Emo) & 0.627 \\
\bottomrule
\end{tabular}
\caption{Automatic evaluation results (SBERT,
\texttt{all-mpnet-base-v2}) for odd-scene-out task. Event = Engaged Events,
Prop = Properties, Emo = Evoked Emotions.}
\label{tab:automatic_results}
\end{table}

The convergence between human consensus (AC1 = 0.761) and
improved computational performance (Acc = 0.693) suggests
that the proposed scene schema captures the
semantic core of situational distinctions.

\subsection{Experiment 2: Human Evaluation of Scene Abstractions}
\label{sec:exp2}

\subsubsection{Setup}

\paragraph{Task design.}
The second evaluation again focuses on the Expression
Profile, this time asking whether humans find the
LLM-generated scene descriptions apt.

For each keyword-sentence pair, participants first generated
their own open-ended interpretation through a self-elicitation
step before viewing any model output, responding to the
following prompts:

\begin{itemize}[itemsep=2pt]
    \item \textbf{\textit{Engaged events}:} ``What happened
    with the [KEYWORD] in the situation? What did they do or
    what occurred to them?''
    \item \textbf{\textit{Generalizable properties}:} ``What
    are the prominent properties of the [KEYWORD] in this
    situation? In your interpretation, what properties stand
    out as most meaningful or relevant in this context?''
    \item \textbf{\textit{Evoked emotions}:} ``Which emotions
    or wishes does the [KEYWORD] evoke in the situation?''
\end{itemize}

This grounds the evaluation in each participant's own
contextual reading rather than in model-generated outputs alone.
They then compared two model-generated
representations (Schema A vs.\ Schema B), selected the one
better aligned with their own interpretation, and rated their
satisfaction on a 5-point Likert scale.
If the rating was less than 5, participants identified reasons for dissatisfaction
from a predefined checklist: \textit{irrelevant information} (not grounded in the situation or unrelated to the keyword), \textit{redundant or verbose} information, \textit{false information} (distorts information from the text), \textit{over-interpretation} (adds information not supported by the text), \textit{lacks information}, \textit{hard to understand} (unclear wording or poorly organized), and \textit{other} (offered a free-text field for elaboration).
The self-elicited prior thus serves as the primary
reference point for assessing alignment throughout the
analysis.

\paragraph{Stimuli and participants.}
The evaluation set comprised 105 target keywords across 120 usage sentence samples drawn from two sources: DWUG-EN \citep{schlechtweg2020semeval}, a dataset designed for lexical semantic change research that provides diverse usage instances of the same word across varying contexts, and COCA \citep{davies2008corpus}.
The same six annotators from Experiment~1,
divided into two groups of three (Group 1 and 2), each evaluated 60 keywords across three dimensions (180 items per participant), yielding three independent judgments per item.\footnote{The study received an IRB exempt determination (STUDY00008122).}

\begin{table}[t]
\centering
\footnotesize
\renewcommand{\arraystretch}{1.1}
\setlength{\tabcolsep}{4pt}
\resizebox{\columnwidth}{!}{%
\begin{tabular}{p{1.4cm}p{1.8cm}p{4.2cm}}
\toprule
\textbf{Category} & \textbf{Relation} & \textbf{Description} \\
\midrule
\multirow{10}{*}{\parbox{1.8cm}{Engaged\\Events\\{\scriptsize (n=10)}}}
 & \texttt{Causes}      & What the event causes \\
 & \texttt{HinderedBy}  & What prevents the event \\
 & \texttt{xReason}     & Why PersonX acts \\
 & \texttt{HasSubEvent} & Steps making up the event \\
 & \texttt{isBefore}    & What usually happens before \\
 & \texttt{isAfter}     & What usually happens after \\
 & \texttt{xIntent}     & What PersonX intends \\
 & \texttt{xNeed}       & What PersonX needs beforehand \\
 & \texttt{xEffect}     & What happens to PersonX \\
 & \texttt{oEffect}     & What happens to others \\
\midrule
\multirow{7}{*}{\parbox{1.8cm}{Gen.\\Properties\\{\scriptsize (n=7)}}}
 & \texttt{ObjectUse}   & What the keyword is used for \\
 & \texttt{HasProperty} & Attribute or quality \\
 & \texttt{MadeUpOf}    & Material or components \\
 & \texttt{AtLocation}  & Where typically found \\
 & \texttt{CapableOf}   & What the keyword can do \\
 & \texttt{Desires}     & What it wants (if animate) \\
 & \texttt{NotDesires}  & What it avoids (if animate) \\
\midrule
\multirow{5}{*}{\parbox{1.8cm}{Evoked\\Emotions\\{\scriptsize (n=5)}}}
 & \texttt{xReact}      & How PersonX feels afterward \\
 & \texttt{oReact}      & How others feel \\
 & \texttt{xWant}       & What PersonX wants next \\
 & \texttt{oWant}       & What others want \\
 & \texttt{xAttr}       & How PersonX is perceived \\
\bottomrule
\end{tabular}%
}
\caption{ATOMIC-2020 relations \citep{hwang2021comet} used in
ATOMIC Profile, consolidated into three categories matching the
evaluation dimensions (Events: $n$=10, Properties: $n$=7,
Emotions: $n$=5).}
\label{tab:atomic_relations}
\end{table}

\paragraph{Baseline.}
We compare Schema A (Scene Profile) against Schema B
(ATOMIC Profile) that operationalizes 22 relations in
ATOMIC-2020 \citep{hwang2021comet}, consolidated into three output fields matching the evaluation dimensions
(Table~\ref{tab:atomic_relations}). Both schema outputs are generated
using \texttt{gpt-4o-mini}. Schema B provides a principled
commonsense baseline: like Schema A, it analyzes a given
situation with respect to a target keyword, but derives
type-level generalizations rather than scene-specific
abstractions, making the comparison a direct test of
instance-level versus type-level semantic representation.

\begin{table*}[t]
\centering
\footnotesize
\renewcommand{\arraystretch}{1.6}
\begin{tabular}{p{2.2cm}p{5.7cm}p{7.0cm}}
\toprule
\multicolumn{3}{l}{%
  \textit{Keyword:} \textbf{whiskey} \quad
  \textit{Situation:} \textit{``The man sat alone at the kitchen table,
  drinking \textbf{whiskey} late at night.''}%
} \\
\midrule
\textbf{Dimension} &
\textbf{Schema A} \newline (Scene Profile) &
\textbf{Schema B} \newline (ATOMIC Profile) \\
\midrule
Engaged Events &
\textbullet\ PersonX drinks it \newline
\textbullet\ It is consumed alone at ObjectY &
{\scriptsize\texttt{Causes:}} whiskey consumption may lead to intoxication \newline
{\scriptsize\texttt{HinderedBy:}} lack of whiskey, being interrupted \newline
{\scriptsize\texttt{HasSubEvent:}} whiskey is poured, whiskey is consumed \newline
{\scriptsize\texttt{isBefore:}} the man finishes drinking \newline
{\scriptsize\texttt{isAfter:}} the man feels effects of alcohol \newline
{\scriptsize\texttt{xEffect:}} the man may feel relaxed or intoxicated \newline
{\scriptsize\texttt{oEffect:}} others may notice his behavior \\
\midrule
Generalizable Properties &
\textbullet\ Often associated with solitude and reflection \newline
\textbullet\ Can signify coping mechanisms during difficult times &
{\scriptsize\texttt{ObjectUse:}} used for drinking, often in social settings \newline
{\scriptsize\texttt{HasProperty:}} alcoholic beverage, can be strong \newline
{\scriptsize\texttt{MadeUpOf:}} distilled grains, water, yeast \newline
{\scriptsize\texttt{AtLocation:}} in a glass, on the kitchen table \newline
{\scriptsize\texttt{CapableOf:}} causing intoxication, providing warmth \\
\midrule
Evoked Emotions &
\textbullet\ Melancholy \newline
{\scriptsize (deeper emotional state tied to drinking alone)} &
{\scriptsize\texttt{xReact:}} the man feels reflective or melancholic \newline
{\scriptsize\texttt{oReact:}} others may feel concerned or indifferent \newline
{\scriptsize\texttt{xWant:}} to escape or forget \newline
{\scriptsize\texttt{oWant:}} to check on the man or leave him alone \\
\bottomrule
\end{tabular}
\caption{Schema A (Scene Profile) vs.\ B (ATOMIC Profile) outputs for \textit{whiskey}, both generated by \texttt{gpt-4o-mini}.}
\label{tab:schema_example}
\end{table*}

\subsubsection{Results}

\paragraph{Overall preference.}
Scene Profiles were preferred in \textbf{86.4\%} of 1,026
valid evaluations (360 Engaged Events, 360 Generalizable
Properties, 306 Evoked Emotions), substantially exceeding
chance ($p < .001$, binomial test for each dimension).
When annotators selected Scene Profiles, they did so with high satisfaction (mean: 4.78 / 4.64 / 4.71 across dimensions, SD $\leq$ 0.78), significantly higher than when ATOMIC Profiles were selected (mean: 3.97 / 4.33 / 4.46, SD $\leq$ 0.93; Mann--Whitney $U$, $p < .01$ for all dimensions). 
Table~\ref{tab:exp2_preference_rates} summarizes results by dimension.

\begin{table}[t]
\centering
\small
\renewcommand{\arraystretch}{1.2}
\setlength{\tabcolsep}{4pt}
\begin{tabular}{lccc}
\toprule
\textbf{Dimension} & \textbf{Pref.} &
\textbf{Scene} &
\textbf{ATOMIC} \\
& \textbf{\%} & \textbf{rating} & \textbf{rating} \\
\midrule
Engaged Events   & 91.4\% & 4.78 $\pm$ 0.59 & 3.97 $\pm$ 0.71 \\
Gen.\ Properties & 89.2\% & 4.64 $\pm$ 0.78 & 4.33 $\pm$ 0.93 \\
Evoked Emotions  & 77.1\% & 4.71 $\pm$ 0.70 & 4.46 $\pm$ 0.91 \\
\midrule
Overall          & 86.4\% & --- & --- \\
\bottomrule
\end{tabular}
\caption{Human preference rates for Scene Profiles over ATOMIC Profiles and mean satisfaction ratings (5.0 scale) when each profile type was selected.}
\label{tab:exp2_preference_rates}
\end{table}

\paragraph{Inter-annotator agreement.}
Gwet's AC1 yielded substantially higher agreement (Table
\ref{tab:exp2_iaa_results}). 
Agreement was consistently higher for Group 1
(AC1: 0.80--0.90, Substantial to Almost Perfect)
than Group 2 (AC1: 0.32--0.70, Fair to Substantial),
reflecting annotator-level variation in evaluation style.

\begin{table}[t]
\centering
\footnotesize
\renewcommand{\arraystretch}{1.15}
\setlength{\tabcolsep}{3pt}
\begin{tabular}{p{1.8cm}p{2.2cm}p{2.0cm}p{1.0cm}}
\toprule
\textbf{Metric} & \textbf{Group 1} & \textbf{Group 2} &
\textbf{Mean} \\
\midrule
\multicolumn{4}{l}{\textit{Engaged Events}} \\
Full Agr.  & 85\%                       & 65\%                      & 75\%    \\
Gwet's AC1      & 0.89 (Alm.\ Perf.)& 0.70 (Subst.)    & 0.80 \\
\midrule
\multicolumn{4}{l}{\textit{Generalizable Properties}} \\
Full Agr.  & 87\%                       & 58\%                      & 72.5\%  \\
Gwet's AC1      & 0.90 (Alm.\ Perf.)& 0.61 (Subst.)    & 0.76 \\
\midrule
\multicolumn{4}{l}{\textit{Evoked Emotions}} \\
Full Agr.  & 76\%                       & 43\%                      & 59.5\%  \\
Gwet's AC1      & 0.80 (Subst.)     & 0.32 (Fair)      & 0.56 \\
\bottomrule
\end{tabular}
\caption{IAA results for Experiment~2. Full Agr.\ = Full Agreement ratio.
Interpretation of Gwet's AC1 follows \citet{landis1977measurement}.}
\label{tab:exp2_iaa_results}
\end{table}

\paragraph{Dimension-specific findings.}
Scene Profile was preferred over ATOMIC Profiles across all three dimensions, with preference rates of 91.4\% for Engaged Events,
89.2\% for Generalizable Properties, and 77.1\% for Evoked Emotions.
Each dimension is examined below.

\textit{Engaged events.} Scene Profile's strength lies in
recovering event participation that goes \textit{beyond the
explicitly stated event structure} of the scene. It correctly
infers that \textit{stab} in ``my stab at this version of
normality'' denotes an attempt rather than a physical act;
that \textit{rat} in ``at the sight of the rat he stiffened
and shrank, trembling and shuddered'' is identified as a
stimulus in a perceptual event (causing PersonX to react with
fear) even though no rat action is described; and that
\textit{raven} in sports score data encodes a competing team
(all three annotators, satisfaction: 5 in each case).
Of the 31 ATOMIC-preferred cases (8.6\%), 25 (81\%) were due to insufficient elaboration: sometimes annotators found Scene Profile's output correct but too sparse. 
For example, for \textit{pin} in \textit{``Cousin James, carrying a baseball bat, pinned Jay to the floor''}, Scene Profile outputs \textit{``PersonX pins PersonY down''}, correctly identifying the core action but with minimal elaboration. ATOMIC Profile outputs \textit{``Cousin James uses the baseball bat to pin Jay; Jay is held down''}, which one annotator found more complete and descriptive. In no case did more than one of annotator prefer ATOMIC Profile for Engaged Events.

\textit{Generalizable properties.} Scene Profile captures
properties activated by the scene rather than given by
encyclopedic defaults.
For \textit{quilt} in \textit{``Miss Silence had fully worked herself up to the determination of going to the quilting''}, Scene Profile outputs \textit{``represents a communal activity involving creativity; associated with social gatherings''}, while ATOMIC Profile outputs \textit{``used for: warmth, decoration, or comfort; made of fabric, often colorful''}.
For \textit{rat} in \textit{``at the sight of the rat
he stiffened and shrank, trembling and shuddered''},
Scene Profile outputs \textit{``can evoke strong fear
responses in individuals''}, while ATOMIC Profile outputs
\textit{``made of: fur, bones, teeth''}.

\begin{table*}[t]
\centering
\small
\setlength{\tabcolsep}{5pt}
\renewcommand{\arraystretch}{1.2}
\begin{tabular}{l r r r r r r r r r}
\toprule
& & \multicolumn{8}{c}{\textit{Reasons for preferring ATOMIC Profile (\% among ATOMIC-preferred cases)}} \\
\cmidrule(lr){3-10}
\textbf{Dimension} &
\parbox[t]{1.2cm}{\raggedleft\textbf{\%\\ATOMIC\\preferred}} &
\parbox[t]{1.2cm}{\raggedleft\textbf{Lacks\\info.}} &
\parbox[t]{1.2cm}{\raggedleft\textbf{Over-\\interp.}} &
\parbox[t]{1.2cm}{\raggedleft\textbf{False\\info.}} &
\parbox[t]{1.2cm}{\raggedleft\textbf{Irrelevant}} &
\parbox[t]{1.2cm}{\raggedleft\textbf{Verbose}} &
\parbox[t]{1.4cm}{\raggedleft\textbf{Hard to\\understand}} &
\parbox[t]{0.8cm}{\raggedleft\textbf{N/A}} &
\parbox[t]{0.8cm}{\raggedleft\textbf{Other}} \\[18pt]
\midrule
Engaged Events   & 8.6\%  & 77\% & ---  & 13\% & ---  & ---  & 6\%  & \cellcolor{black!10} & 16\% \\
Gen. Properties & 10.8\% & 49\% & 28\% & 15\% & 10\% & 8\%  & 15\% & \cellcolor{black!10} & 5\%  \\
Evoked Emotions  & 22.9\% & 41\% & 36\% & 13\% & 3\%  & 3\%  & 3\%  & 24\% & 16\% \\
\bottomrule
\end{tabular}
\caption{Cases where ATOMIC Profile was preferred over
Scene Profile, with breakdown of annotators' reasons
(percentages among ATOMIC-preferred cases per dimension).
Each annotator could select multiple reasons, so
percentages within a row may sum to more than 100\%.
\textit{Irrelevant}: not grounded in the situation or
unrelated to the keyword; \textit{Over-interp.}: adds
information not supported by the text; \textit{False
info.}: distorts information from the text;
\textit{Verbose}: redundant or unnecessarily verbose;
\textit{Hard to understand}: unclear wording or poorly
organized; \textit{N/A}: Scene Profile produced no evoked
emotion output for this usage instance. This category
applies only to Evoked Emotions; grayed cells indicate
it is not applicable to the other two dimensions.}
\label{tab:failure_breakdown}
\end{table*}

\textit{Evoked emotions.}
Scene Profile was preferred less often for Evoked Emotions (77.1\%) than for the
other two dimensions,
which may reflect a structural
difference between the two profiles.
Scene Profile is designed to capture the emotion evoked in the reader by the target expression, committing to one interpretive stance. ATOMIC
Profile, by contrast, enumerates the reactions and desires of multiple participants within the scene (\texttt{xReact}, \texttt{oReact}, \texttt{xWant},
\texttt{oWant}), which may align more readily with annotators who foregrounded different participants.
For \textit{beer} in \textit{``he had been training for
months and habitually refusing my offers to bring over
beer on the weekends''}, annotators described the evoked
emotion as ``wish to socialize'', ``beer is not wanted
here'', and ``socialization'', each reflecting a
different participant's perspective. Scene Profile output
\textit{``Disappointment''}, aligning with one reading
but not others, while ATOMIC Profile output
\textit{``wants to: avoid distractions from training;
others want to: socialize and enjoy drinks together''},
accommodating multiple perspectives. All three annotators
preferred ATOMIC Profile in this case.

\paragraph{Failure case analysis.}
Table~\ref{tab:failure_breakdown} summarizes failure modes
across dimensions.
Scene Profile's failures largely fall into two patterns. In \textit{lacks-info} cases, Scene Profile correctly identified the scene role but did not elaborate it sufficiently: for \textit{wine} in \textit{``we sit
outside at a table with a checked cloth and drink wine in the sun''} (Engaged Events), Scene Profile output only \textit{``PersonX and PersonY drink it''}, which annotators found too sparse.
In \textit{over-interpretation} cases, Scene Profile inferred a property not warranted by the scene: for \textit{lavender} in \textit{``it had like a purple satin ruffle skirt, maybe lavender''} (Generalizable Properties), Scene Profile output \textit{``may evoke a sense of calm or tranquility''}, while an annotator's own response was simply \textit{``it's the color of the skirt''}.
For Evoked Emotions, Scene Profile sometimes produced
\texttt{None} when it identified no specific emotions
evoked by the target expression. Annotators who disagreed
with this judgment found the output unsatisfying (24\% of
ATOMIC-preferred Emotion cases; see N/A column in
Table~\ref{tab:failure_breakdown}).
The over-interpretation rate was consistently higher when annotators reported ``nothing specific'' in their self-elicited response (Property: 33.3\% vs.\ 10.1\%; Emotion: 20.7\% vs.\ 7.9\%), 
suggesting that Scene Profile over-interprets most when the scene itself provides insufficient grounding for a specific inference.

\paragraph{Qualitative feedback.}
At the end of the experiment, annotators provided
free-text assessments of each profile's overall
strengths and weaknesses (see Table~\ref{tab:qualitative}
in Appendix C for full responses). Scene Profile was consistently
characterized as more concise and contextually accurate,
though some annotators noted a tendency to over-interpret
or to summarize the scene rather than focus on the target
word. ATOMIC Profile was described as more thorough and
complete, but frequently criticized for being verbose,
redundant, and including information irrelevant to the
specific usage context.

\section{Discussion}
\label{sec:discussion}

\subsection{What Scene Representations Capture}

The results of Experiment~1 provide converging evidence from
two independent evaluation tracks that scene-level distinctions
constitute a meaningful unit among human participants.
They identified outlier situations of target word instances
with 82.37\% accuracy, far above the 20\% chance level,
and with substantial inter-annotator consensus
(Gwet's AC1 = 0.761). This suggests that scene-level
distinctions reflect shared perceptual-interpretive structure
rather than task-specific conventions.

The computational results mirror this pattern. Integrating
scene features raises odd-scene-out accuracy from 0.575 to
0.693 (+11.8 percentage points), and the scene-only condition alone
outperforms the text baseline (0.627 vs.\ 0.575). Scene
abstractions derived entirely from LLM-generated structured
representations, with no access to the original sentence,
carry sufficient semantic weight to distinguish situational
outliers.
This suggests that scene features may reflect
a distinct layer of situational interpretation
rather than a
restatement of surface lexical content.

Among individual scene components in the odd-scene-out task,
\textit{Generalizable Properties} emerges as the most
discriminative feature (0.661), outperforming both Engaged
Events (0.609) and Evoked Emotions (0.562).
This pattern may reflect the nature of the keywords chosen in COCA-Scenes, which are predominantly concrete nouns whose situational configurations are particularly well distinguished by property-level attributes.

\subsection{Scene-Grounding as a Mechanism of Alignment}

As reported in Section~\ref{sec:exp2}, Scene Profiles were
preferred in 86.4\% of evaluations across all three
dimensions. 
This advantage stems not from general output quality but from
a representational difference:
Scene Profile conditions its abstraction on the particular scene instance. ATOMIC Profile, by contrast, is organized around relational schemas that describe what typically holds for a keyword in general; as shown in Table~\ref{tab:schema_example}, this can result in outputs that apply broadly across contexts rather than capturing what is specific to the scene at hand.

\textit{Lacks-info} was the most common reason for
preferring ATOMIC Profile in Engaged Events (77\% of
ATOMIC-preferred cases), while \textit{over-interpretation}
was more prominent in Generalizable Properties (28\%) and
Evoked Emotions (36\%).
Over-interpretation rates were consistently higher when
annotators wrote ``nothing specific'' in their
self-elicitation response (Property: 33.3\%
vs.\ 10.1\%; Emotion: 20.7\% vs.\ 7.9\%), 
suggesting that Scene Profile is prone to overreaching
or hallucinating when the usage instance provides
little scene-relevant information to draw from.

\section{Conclusion}
\label{sec:conclusion}

We introduced Scene Abstraction, a computational approach
that constructs structured scene representations
$\mathcal{S}(u, x)$ for word usage instances.
Each scene comprises a Contextual Scene $\mathcal{C}$ capturing
the broader situational interpretation and an Expression
Profile $\mathcal{E}$ capturing the scene-grounded meaning of
a target expression, operationalized via few-shot prompting
of an LLM.

Two experiments provide complementary support for the
framework.
Experiment~1 shows that scene-level distinctions are
reliably identifiable across human observers (82.37\%
accuracy, AC1 = 0.761) and that scene-based representations
outperform raw-text embeddings on this task (+11.8 pp).
Experiment~2 shows that
LLM-generated scene profiles align more
closely with human scene-grounded word meaning than
ATOMIC-based commonsense profiles (86.4\% preference).

Together, these results suggest that scene representations reflect aspects of how humans interpret words in situated contexts. The three evaluation dimensions (what event a word engages in, what properties it bears, and what emotion it evokes) show different patterns of human agreement, pointing to the complexity of scene-grounded interpretation.
Scene Abstraction offers one way to make aspects of this interpretation explicit and computationally accessible, complementing existing distributional and knowledge-based approaches.

Several directions emerge naturally from this work.
Extending the framework to metaphorical and non-literal language is a natural next step, as metaphor often arises from structural and property-level similarities across situations, dimensions that scene representations capture explicitly.
Aggregating scene representations across usage instances to study a type-level semantics of a word is another direction worth exploring.

\subsection*{Limitations}
The framework has several limitations worth noting.
Scene abstractions are generated by a prompted LLM and may
inherit systematic biases or hallucinated inferences from
the underlying model. The affective and interpretive
dimensions of scenes (inferred properties and emotions,
atmosphere, evoked emotions) are understood as subjective
inferences from situational context rather than fixed,
objective semantic properties, and the current evaluation
design does not fully control for this subjectivity. We
make no claim that the scene representations produced by
this framework correspond directly to cognitive
representations in the psychological sense; the framework
is a representational and analytical tool, not a model of
mental processing. Additionally, the framework has been
validated primarily on English-language fiction;
generalizability across languages, genres, and cultural
contexts remains to be assessed.

\bibliography{custom}

\clearpage
\appendix
\section*{Appendix}

\noindent\begin{minipage}{\textwidth}

\subsection*{A\quad COCA-Scenes Keywords}
\label{app:keywords}
Table~\ref{tab:keywords} lists the 26 keywords used in the
COCA-Scenes dataset.

\vspace{10pt}
\centering
\begin{tabular}{lllllll}
\toprule
bathroom & beer & chicken & cigar & cigarette & coffee & crow \\
eagle & elevator & fire & knife & microwave & oven & owl \\
pistol & raccoon & rat & raven & rifle & squirrel & sword \\
tea & turkey & vulture & water & wine & & \\
\bottomrule
\end{tabular}
\captionof{table}{The 26 keywords in the COCA-Scenes dataset.}
\label{tab:keywords}

\vspace{18pt}
\raggedright

\subsection*{B\quad Prompt Instruction}
\label{app:prompt}

Figure~\ref{fig:prompt_instruction} shows the full prompt
instruction used for scene abstraction.

\vspace{10pt}
\centering
\begin{tcolorbox}[
    enhanced,
    sharp corners,
    colback=gray!5,
    colframe=black,
    boxrule=0.5pt,
    width=0.95\textwidth,
    center,
    title=\small\textbf{Prompt: Scene Abstraction Instructions},
    fonttitle=\sffamily\bfseries
]
Given a sentence and a target keyword, extract a structured
scene abstraction and format your response as a bullet point
list, covering the following fields:
\begin{description}[leftmargin=10pt, itemsep=10pt]
    \item[1. Events:] \hfill \\ \vspace{-1.2\baselineskip}
    \begin{itemize}[label={-}, nosep, leftmargin=15pt, topsep=0pt]
        \item List only what happened in COMET-ATOMIC style (e.g., ``PersonX opens ObjectY'').
        \item Do not include properties, emotions, or interpretations.
        \item Do not include definitions or static facts.
    \end{itemize}
    \item[2. Entities:] \hfill \\ \vspace{-1.2\baselineskip}
    \begin{itemize}[label={-}, nosep, leftmargin=15pt, topsep=0pt]
        \item List each entity with its surface mention
        (e.g., ``PersonX (she)'').
        \item For each entity, include:
        \begin{itemize}[label={$\circ$}, nosep, leftmargin=10pt]
            \item Role(s) with associated Frame(s)
            \item Property: context-specific traits with brief explanations
            \item Emotion: inferred emotional states with explanations
        \end{itemize}
    \end{itemize}
    \item[3. Setting:] \hfill \\ \vspace{-1.2\baselineskip}
    \begin{itemize}[label={-}, nosep, leftmargin=15pt, topsep=0pt]
        \item Describe Place, Time, and Atmosphere if inferable.
        \item Use ``unspecified'' when not identifiable.
    \end{itemize}
    \item[4. Expression Profile:] \hfill \\ \vspace{-1.2\baselineskip}
    \begin{itemize}[label={-}, nosep, leftmargin=15pt, topsep=0pt]
        \item Keyword: the target linguistic expression
        (with its assigned label)
        \item Engaged Events: COMET-ATOMIC style events
        involving the keyword
        \item Generalizable Properties: abstract traits
        implied by the context
    \end{itemize}
\end{description}
\medskip
\hrule
\medskip
\textbf{Constraints:}
\begin{itemize}[leftmargin=15pt, nosep]
    \item All entity labels must use X/Y/Z suffixes only
    (e.g., PersonX).
    \item Format must be a bullet-style list without extra explanation.
\end{itemize}
\end{tcolorbox}
\captionof{figure}{The prompt instruction for the scene abstraction process.}
\label{fig:prompt_instruction}

\end{minipage}

\clearpage

\begin{table*}[!ht]
\centering
\subsection*{C\quad Annotator Qualitative Feedback}
\label{app:qualitative}
\vspace{10pt}
\small
\renewcommand{\arraystretch}{1.3}
\setlength{\tabcolsep}{4pt}
\begin{tabular}{p{0.4cm} p{3.4cm} p{3.4cm} p{3.4cm} p{3.4cm}}
\toprule
\textbf{ID} & \textbf{Strengths A} & \textbf{Weaknesses A} &
\textbf{Strengths B} & \textbf{Weaknesses B} \\
\midrule
1 &
Far more concise; far more accurate; far better at understanding the context behind word meanings; rarely misinterprets a scene or adds irrelevant information &
Occasionally over interprets a scene, finding unnecessary symbology; occasionally inaccurate; struggles with identifying the impact of JUST the word (resorts to summarizing the scene); exaggerates &
I guess it knows a lot... it establishes a LOT of possibilities and contextual information, such as what the word does in other scenarios, what it's made up of, etc.; generally knows the context behind word meanings &
Struggles with identifying the impact of JUST the word (resorts to summarizing the scene); misinterprets scenes often (struggles with identifying the proper use of the word when it's used as an idiom or other such expression); horribly redundant; often over interprets a scene; provides far too much information; formatting is confusing and unnecessary; exaggerates \\
\addlinespace
2 &
Very concise; mostly easy to understand &
Often very broad; sometimes written very scientifically, i.e., unnecessarily complicated &
Usually covers the whole range of the situation; very specific to the situation; is written in a way that is easy to understand (like everyday speech) &
Often takes things too literally; almost always adds unnecessary information that does not have much to do with the specific situation at hand \\
\addlinespace
3 &
I liked the sentence structure, and it was more concise which made it easier to read. There wasn't as much irrelevant information. &
I didn't always like the structure of Answer A when talking about the emotions that a word evoked. &
When talking about the emotions that certain words or context evoked, it had good sentence structure sometimes. &
Generally was not concise and had more irrelevant information. Especially when talking about the properties, the sentence structure had duplicate words and took more comprehension to understand. \\
\addlinespace
4 &
It was better at figuring out the meaning of the word in context than B; A generally did a better job at only giving information pertinent to the context; I was generally pretty satisfied with its responses &
N/A &
Giving the idea in multiple ways is helpful &
Sometimes it did so to an extreme degree and made it difficult to read; the second one overextrapolated from the scene way more than the first \\
\addlinespace
5 &
Very concise and direct, to the point; good at identifying the sense of the word, it didn't mix up autumn (color) with autumn (season); captures the emotions and thoughts provoked by the word well &
Sometimes very vague with describing what happens to the object in the scene &
Very descriptive and complete with its responses, captures the entire scene in great detail &
Very verbose and sometimes redundant; sometimes misinterprets the sense of the word, mixes autumn (color) with autumn (season); reads too much into the scene, doesn't generalize well (typically found in..., but it is specific to the scene only) \\
\addlinespace
6 &
Tend to give more direct answers that relate to the context of the sentence; more direct and answers the questions more clearly &
N/A &
More thorough but adds extra information &
Tend to give a lot more irrelevant and verbose information and often over-analyzes the word; often ones that are related to the words at hand but are not relevant to the context \\
\bottomrule
\end{tabular}
\caption{Annotator free-text assessments of \textbf{Scene
Profile (A)} and \textbf{ATOMIC Profile (B)}, collected at the end
of Experiment~2. Annotators 1--3 form Group~1 and
Annotators 4--6 form Group~2.}
\label{tab:qualitative}
\end{table*}

\end{document}